\documentclass[tinyml]{acmart}

\AtBeginDocument{%
  \providecommand\BibTeX{{%
    \normalfont B\kern-0.5em{\scshape i\kern-0.25em b}\kern-0.8em\TeX}}}
    
\setcopyright{rightsretained}
\copyrightyear{2022}
\acmYear{2022}

\usepackage[utf8]{inputenc}
\usepackage{subfigure}
\usepackage{amsmath,amsfonts}
\usepackage{algorithm2e}
\usepackage{multirow}
\usepackage{float}
\usepackage{booktabs}
\usepackage{dashrule}
\usepackage{xcolor}
\usepackage{multirow}

\begin{document}

\title{A Fast Network Exploration Strategy to Profile Low Energy Consumption for Keyword Spotting}

\author{Arnab Neelim Mazumder}
\email{arnabm1@umbc.edu}
\affiliation{%
  \institution{University of Maryland Baltimore County}
  \streetaddress{1000 Hilltop Circle}
  \city{Baltimore}
  \state{Maryland}
  \country{USA}
  \postcode{21250}
}

\author{Tinoosh Mohsenin}
\email{tinoosh@umbc.edu}
\affiliation{%
  \institution{University of Maryland Baltimore County}
  \streetaddress{1000 Hilltop Circle}
  \city{Baltimore}
  \state{Maryland}
  \country{USA}
  \postcode{21250}
}

\begin{abstract}
Keyword Spotting nowadays is an integral part of speech-oriented user interaction targeted for smart devices. To this extent, neural networks are extensively used for their flexibility and high accuracy. However, coming up with a suitable configuration for both accuracy requirements and hardware deployment is a challenge. We propose a regression-based network exploration technique that considers the scaling of the network filters ($s$) and quantization ($q$) of the network layers, leading to a friendly and energy-efficient configuration for FPGA hardware implementation. We experiment with different combinations of $\mathcal{NN}\scriptstyle\langle q,\,s\rangle \displaystyle$ on the FPGA to profile the energy consumption of the deployed network so that the user can choose the most energy-efficient network configuration promptly. Our accelerator design is deployed on the Xilinx AC 701 platform and has at least
2.1$\times$ and 4$\times$ improvements on energy and energy efficiency results, respectively, compared to recent hardware implementations for keyword spotting.
\end{abstract}


\maketitle

\section{Introduction}
\label{sec:intro}

The recent advancements in deep learning have led to significant breakthroughs in computer vision, data analytics, text authentication, speech recognition, etc. The application requirements in these fields have forced researchers to concentrate on making neural networks more flexible and accurate to a degree where they can surpass human-level accuracies. Speech recognition is one such domain where the deep learning techniques have allowed the introduction and evolution of voice assistant devices like Siri, Alexa, Amazon Echo, etc. However, speech recognition requires devices to be always on, which burdens battery life. Additionally, speech recognition tasks utilize cloud-based solutions where communication and latency are an issue. One solution to these problems is to apply keyword spotting (KWS) on resource-constrained devices. KWS helps detect specific keywords such as ‘Hey Siri, ‘Ok Google,’ etc., and can trigger the voice assistant devices to become active. This avoids the issue of the systems being always-on and thus helps in saving battery life. However, the KWS networks themselves have to be always on to look for specific keywords. Hence, KWS networks need to be accurate and consume low power during execution. With the ongoing developments in making deep neural networks (DNNs) compressed, sparse, and flexible during inference, KWS has gained a lot of attention within the research community in recent years \cite{chen2015locally, arik2017convolutional}.

The requirement of a low power envelope means that resource-constrained devices like low-cost field-programmable gate arrays (FPGAs), microcontrollers, and similar edge devices should be utilized to accelerate the KWS networks \cite{mazumder2021survey}. This further strengthens the fact that neural network frameworks targeted for KWS need to be compressed and efficiently accelerated so that they can fit on tiny edge devices with stringent constraints on power and memory. In addition to this, the choice of the network parameters concerning hardware implementation is always a challenge. So, we need a mechanism to select the network parameters that lead to the most accurate software architecture while considering the hardware deployment constraints. Thus, we define KWS as a tinyml application and aim to find a solution for selecting suitable networks parameters for deployment on commercial off-the-shelf FPGAs or commodity microcontrollers. To achieve this, we propose a regression-based neural network exploration technique that finds the best software parameters in a hardware-aware fashion. The notable contributions of this work include:

\begin{itemize}
    \item Introducing a systematic approach based on experimental and analytical methods to profile energy consumption of DNNs on FPGAs for KWS.
    \item Presenting two polynomial regression setups that predict the accuracy of quantized and scaled DNNs for KWS and estimate their energy consumption on hardware.
    \item Developing parameterized and scalable hardware for any number of processing engines (PEs) and precision levels for implementing the same task with comparable accuracy on a tiny low-power, low-cost FPGA. 
\end{itemize}

\section{Related Work}
\label{sec:related_work}

KWS has been traditionally implemented through convolutional neural networks (CNNs) for a variety of platforms including mobile devices \cite{rybakov2020streaming}, microcontrollers \cite{banbury2021micronets}, and FPGAs \cite{shan202014}. However, alternate networks using bidirectional long short-term memory networks (LSTMs) \cite{zeng2019effective} and convolutional recurrent neural networks (CRNN) \cite{khursheed2021tiny} have also emerged as viable solutions over the years. One of the bottlenecks of KWS is to process the audio waveforms, and utilizing Mel-frequency cepstrum (MFCC) for audio to frame conversion has been a popular technique \cite{shan202014}. But recently, sinc convolutions have been introduced that provide an end-to-end solution for processing raw audios \cite{mittermaier2020small}. Furthermore, compression of the networks to their low bitwidth fixed point counterparts is another aspect of KWS that is imperative for efficient hardware deployment. The developments in the domain of quantization include binarized (BNNs) \cite{hubara2016binarized} and Ternarized neural networks (TNNs) \cite{alemdar2017ternary} which have paved the way to compress DNNs to extreme quantization levels in an accuracy-aware style. However, network exploration for KWS is still one of the most fundamental approaches that allow the selection of the best configuration for KWS workloads. Keeping in line with this, there have been neural architecture search (NAS) techniques for KWS which focus on optimizing the network parameters for optimal accuracy in a cell-based manner \cite{mo2020neural}. However, this does not approach the network exploration problem from a hardware deployment point of view. Hardware-aware neural architecture search has been in the literature for quite some time where the concentration is mainly towards efficient co-design of software, and hardware \cite{abdelfattah2020best}. More recently, a regression-based fast NAS setup was explored in \cite{hosseini2021afast, hosseini2021qsnas} for vision applications that aim to optimize energy for FPGA representations of the configurations. This work extends the idea developed in \cite{hosseini2021qsnas} and uses the regression-based NAS strategy to find a fast and suitable solution that will lead to low energy consumption and efficient hardware design.

\section{Background and Problem Formulation}
\label{sec:problem_form}

CNNs utilize matrices or tensors and assign a degree of importance to various parts of the spatial field, known as feature extraction. Allocating importance is the process of using learnable parameters (weights and biases) to help the framework identify contrasting features. Previously, machine learning algorithms required hand-crafted features to be fed to the algorithms to create inference architectures. This is time-consuming and requires an intricate understanding of spatial features. CNNs do not depend on these primitive techniques. Instead, it utilizes the filters in the network to learn these characteristics. In many ways, a CNN network is a replica of the neurons in a human brain where specific parts of the receptive field get triggered upon the visual reception of an image. However, this also makes CNNs very compute-intensive and challenging to deploy on resource-constrained devices.

The major factors that influence the power consumption of a deployed CNN can be listed as (1) scaling of the filters, (2) quantization of the weights and data, (3) resolution of the inputs, (4) depth of the network, (5) sparsity of the network, and (6) interconnection between layers. Thus, coming up with an optimal solution that includes the best features of all six of these is a challenge and is out of scope for this work. Therefore, we focus on the two most influential factors contributing to the energy consumption of deployed CNNs, namely scaling the filters and quantization of the weights and data.

The vanilla 2D convolution is given by Equation \ref{eq:conv_output_comp} where, $I$, $O$, $W$, $C$, and $F$ correspond to input, output, weights, output channels and input channels respectively. The $S$, in this case, represents the kernel strides.

\begin{equation}
\begin{aligned}
\mathcal{O}_{m,n}\scriptstyle \displaystyle = \sum_{\mathcal{C}=1}^{\mathcal{C}_{in}} \scriptstyle \langle \sum_{\mathcal{F}=1}^{\mathcal{F}} \langle \mathcal{I}_{{F+mS},C} * \mathcal{W}_{n,C,F} \rangle \rangle \displaystyle for\hspace{1ex} \scriptstyle m = 0..\mathcal{N}-1, n = 1..\mathcal{C}_{out} \displaystyle
\label{eq:conv_output_comp}
\end{aligned}
\end{equation}

We introduce the scaling variable $(s)$ and quantization variable $q$ to alter the number of filters and quantization in each convolution layer inside the network, respectively. This results in several network configurations with quantization level $q$ and scaling $s$. Hence, we frame a problem in the following mold:

\begin{equation} 
\footnotesize
\begin{aligned}
minimize\hspace{1ex} \mathcal{NN}\langle Energy\rangle for\hspace{1ex}config \in\{P, M, q, s\}
\label{eq:opt_problem}
\end{aligned}
\end{equation}

This minimizes the energy generated for inference of these models provided that the target accuracy is maintained where $P$ and $M$ correspond to the number of processing engines and multipliers of the hardware. We also consider MFCC spectrograms of the audio signals, albeit this does not affect the formulation described in Equation \ref{eq:opt_problem}. This, however, allows us to set a benchmark input resolution for our workloads. Finally, the above formulations also enable us to utilize approximators to infer the pattern of energy on FPGA and solve the problem through using the following steps:

\begin{enumerate}
\item Use a baseline DNN network to learn the trend for accuracy on a limited span of $q$ and $s$. 
\item Fit an approximator for step (1) through regression.
\item Deploy a handful of the baseline DNN configurations for $q$ and $s$ on the FPGA to learn the energy trend.
\item Fit another approximator that predicts the energy consumption behavior on FPGA through regression.
\item Perform final optimization w.r.t. the two approximators thus developed to choose the best configuration that provides the best trade-off.
\end{enumerate}

\begin{figure}
\begin{center}
\includegraphics[width=\linewidth]{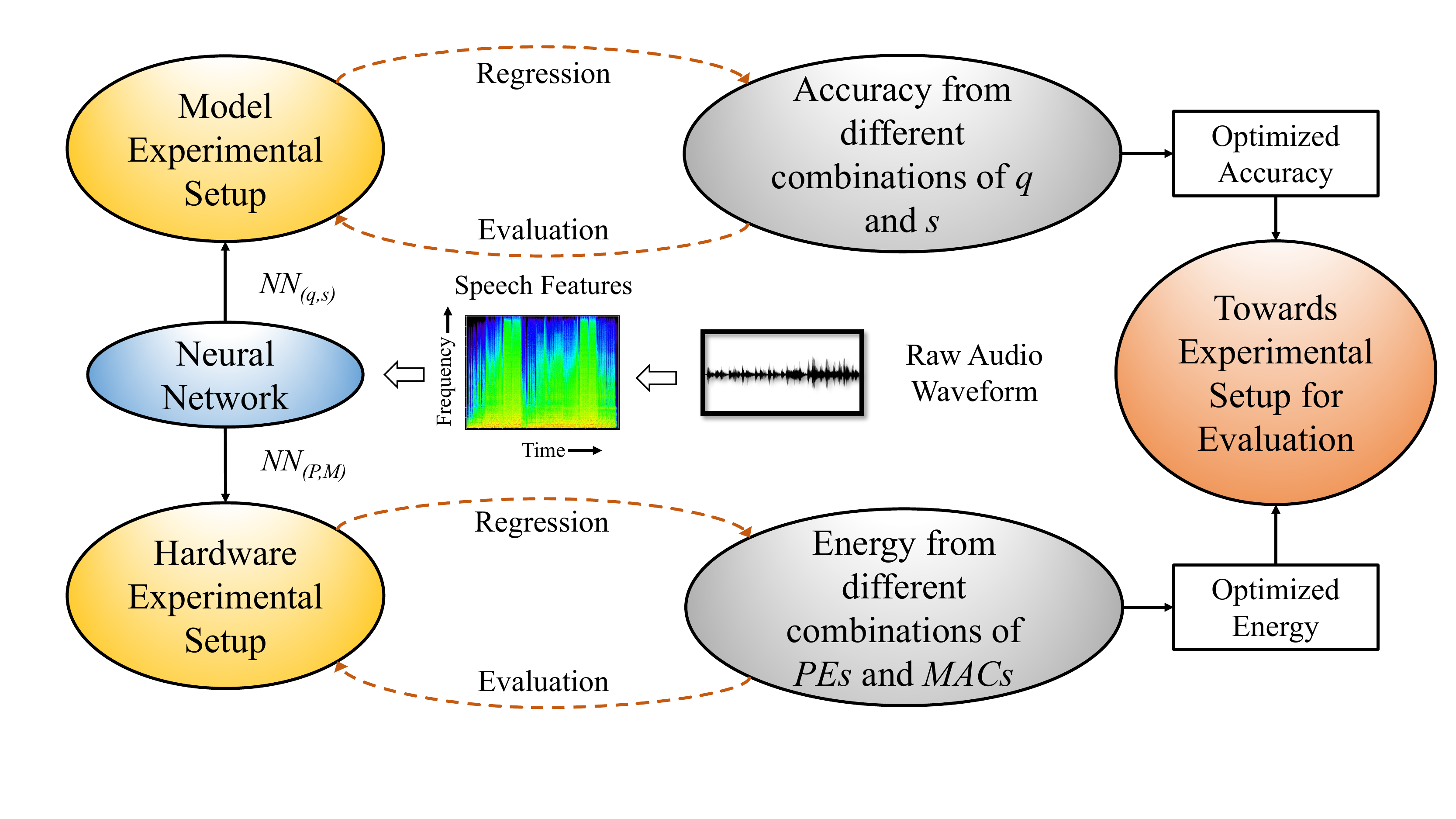}
\end{center}
\vspace{-3ex}
\caption{A high level overview of the required steps for experimental and analytical approach of our proposed methodology for keyword spotting.}
\vspace{-1ex}
\label{fig:overview}
\end{figure}

\begin{table}[]
\centering
\caption{Network Architecture Used for the Experiments and Model Analytics corresponding to $q$ and $s$}
\label{tab:network}
\begin{tabular}{c|ccc}
\hline
 & \multicolumn{3}{c}{Google Speech Commands} \\ \hline
Layer & \multicolumn{1}{c|}{Kernel Shape} & \multicolumn{1}{c|}{\#Filters} & Stride \\
Conv2D & \multicolumn{1}{c|}{3$\times$3} & \multicolumn{1}{c|}{$64s$} & 1 \\
Maxpool2D & \multicolumn{1}{c|}{2$\times$2} & \multicolumn{1}{c|}{$64s$} & 2 \\
Conv2D & \multicolumn{1}{c|}{3$\times$3} & \multicolumn{1}{c|}{$32s$} & 1 \\
Maxpool2D & \multicolumn{1}{c|}{2$\times$2} & \multicolumn{1}{c|}{$32s$} & 2 \\
Conv2D & \multicolumn{1}{c|}{3$\times$3} & \multicolumn{1}{c|}{$32s$} & 1 \\
Maxpool2D & \multicolumn{1}{c|}{2$\times$2} & \multicolumn{1}{c|}{$32s$} & 2 \\
Fully Connected & \multicolumn{1}{c|}{-} & \multicolumn{1}{c|}{$64s$} & - \\
Fully Connected & \multicolumn{1}{c|}{-} & \multicolumn{1}{c|}{\#output} & - \\ \hline
\begin{tabular}[c]{@{}c@{}}Total Computations\\ (Millions)\end{tabular} & \multicolumn{3}{c}{3.06~$s^2$} \\ \hline
\begin{tabular}[c]{@{}c@{}}Model Size \\ (KB)\end{tabular} & \multicolumn{3}{c}{4.20~$qs^2$} \\ \hline
\begin{tabular}[c]{@{}c@{}}Largest Feature\\ Map (KB)\end{tabular} & \multicolumn{3}{c}{3.70~$qs$} \\ \hline
\end{tabular}
\end{table}

Figure \ref{fig:overview} gives an overview of our technique. Finally, with regards to the network described in Table \ref{tab:network}, we can make the following inferences in Equation \ref{eq:fmapsize}:

\begin{equation} 
\footnotesize
\begin{aligned}
% \begin{split}
\mathcal{NN}\scriptstyle\langle size\rangle \displaystyle\ &\propto &&q \,\,.\,\, s^2 \\
\mathcal{NN}\scriptstyle\langle largest\_fmap\rangle \displaystyle\ &\propto &&q \,\,.\,\, s \\
\mathcal{NN}\scriptstyle\langle computations\rangle \displaystyle\ &\propto &&\;\;\;\;\:\,s^2 \\
\mathcal{NN}\scriptstyle\langle Mult\_Op\_Cost\rangle \displaystyle\ &\propto &&q^2 .\: s^2 \\
\mathcal{NN}\scriptstyle\langle Add\_Op\_Cost\rangle \displaystyle\ &\propto &&q \,\,.\,\, s^2 \\
\end{aligned}
\label{eq:fmapsize}
\end{equation}

\section{Dataset Description}
\label{sec:dataset}
To provide proof of concept for our approach, we train the neural network described in Table \ref{tab:network} for keyword spotting on the Google Speech Command dataset \cite{warden2018speech}.
This architecture utilizes convolution-max-pooling pairs to isolate feature information followed by fully connected layers to classify a variety of keywords. The dataset contains 30 different classes of audio that convey information of various keywords like up, down, right, left, etc., in different accents. The length of the audios varies from 0.5s to 1s for all classes. We transform the raw audio signals belonging to the classes into their frequency domain representation with the help of the MFCC procedure as shown in Figure \ref{fig:overview}. Even though the network architecture can take audio signals in their raw form as input, it would be computationally expensive to process the 1s long audio file directly. Hence, the audio inputs go through MFCC decomposition. In order to facilitate the MFCC process, we consider audio samples to be 1s long, with the number of samples equaling 22050. We use the librosa \cite{mcfee2015librosa} signal processing library to generate the audio spectrums where we consider the default values of MFCC coefficients, MFCC interval, and hop length provided by the library itself. This results in the audio spectrums having the final shape of 44$\times$13 where 44 corresponds to the sample numbers after windowing and 13 corresponds to the number of MFCC coefficients.

\section{Accuracy Regression}
\label{sec:acc_reg}

\begin{figure} 
\begin{center}
\includegraphics[width=\linewidth]{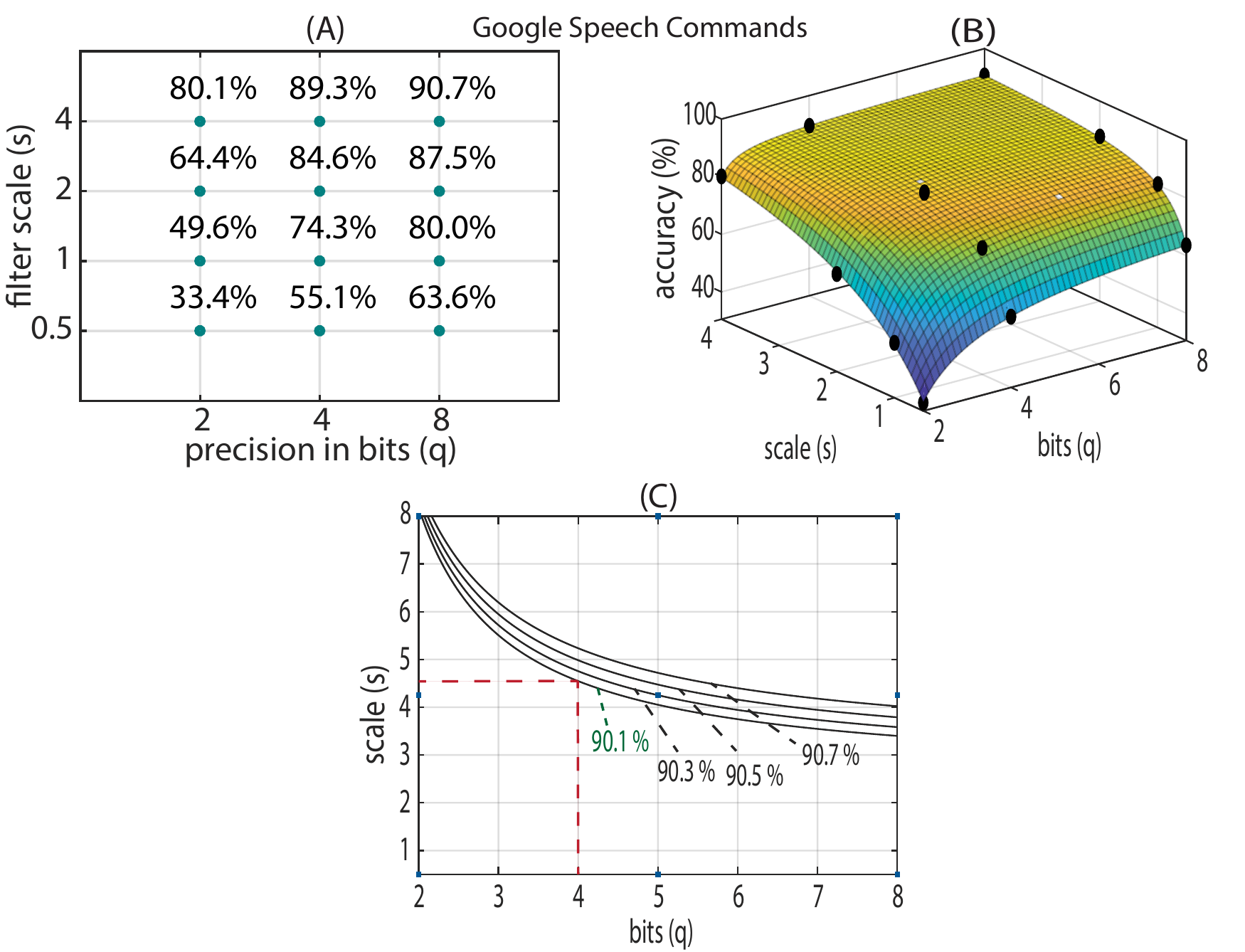}
\end{center}
\vspace{-3ex}
\caption{This figure illustrates the accuracy modeling setup based on empirical analysis for Google speech commands. Different experiences for a span of $q$ and $s$ lead to the surface plots for deciding the unknown values of the coefficients in the approximator Equation \ref{eq:KS}. (A) represents the accuracy experienced from training different $\mathcal{NN}$ configurations, (B) illustrates the surface plot indicating the fit for the custom equation, and (C) demonstrates a better visualization of the relationship between accuracy, scaling, and precision through contours of the applied Equation \ref{eq:KS}.}
\vspace{-3ex}
\label{fig:accuracy_modeling}
\end{figure}

 We train all models for 100 epochs and use Adam optimizer for the experiments with an adaptable learning rate that reduces by a factor of 0.1 at every 33 epochs. With the experimental results from different configurations of scale and quantization, we postulate that by developing a custom equation incorporating the least-square error method represented in Equation \ref{eq:KS}, we can approximate the accuracy level for the DNN with a certain degree of confidence. The degree of confidence comes from the accuracy of the fit for the data points where $\hat{A_i}$ are constant parameters that are learned based on the fit. The data points used to solve the custom equation, its corresponding fitted surface, and accuracy contours are depicted in Figure \ref{fig:accuracy_modeling}. The RMSE for the fitted polynomial is 0.9 indicating a good fit for the custom equation applied. Finally, for the application of keyword spotting, we expect very high accuracy and low latency during operation. Hence, we only highlight the accuracy levels in the contour plot, which are above 90\%.

\begin{equation} 
\footnotesize
\begin{aligned}
% \begin{split}
{Accuracy(\mathcal{NN}\scriptstyle\langle q,\,s\rangle \displaystyle)} \approx \frac{\hat{A_6}.q.s\;+\,\hat{A_5}.s\;\,+\,\hat{A_4}.q\;\;\;+\;\hat{A_3}}{\;\;\;\;\;q.s\;+\hat{A_2}.s\;+\hat{A_1}.q\;+\,\hat{A_0}}
% \end{split}
\end{aligned}
\label{eq:KS}
\end{equation}

\section{Accelerator Design}
\label{sec:hardware}

The primary design goals for the accelerator were to ensure efficient parallel processing for energy and latency boost. The design demonstrated in Figure \ref{fig:top_hard} can be configured for any number of filters, layers, and PEs. Additionally, the design has the flexibility to represent any precision level for data and weights starting from 2-bit up to a 32-bit fixed point. The hardware experiments are performed on the AC~701 evaluation platform, which incorporates Xilinx Artix–7 XC7A200TFBG676-2 FPGA (13.14Mb (= 365$\times$36Kb) on-chip BRAMs) with the toggling rate for switching set to 100\%. This device is in the same range as low-cost microcontrollers in performance and memory. Thus, it is suitable for accelerating tiny neural networks with a size below 1~MB. The main blocks of our implementation are as follows:

\begin{figure}[t] 
\begin{center}
\includegraphics[width=\linewidth]{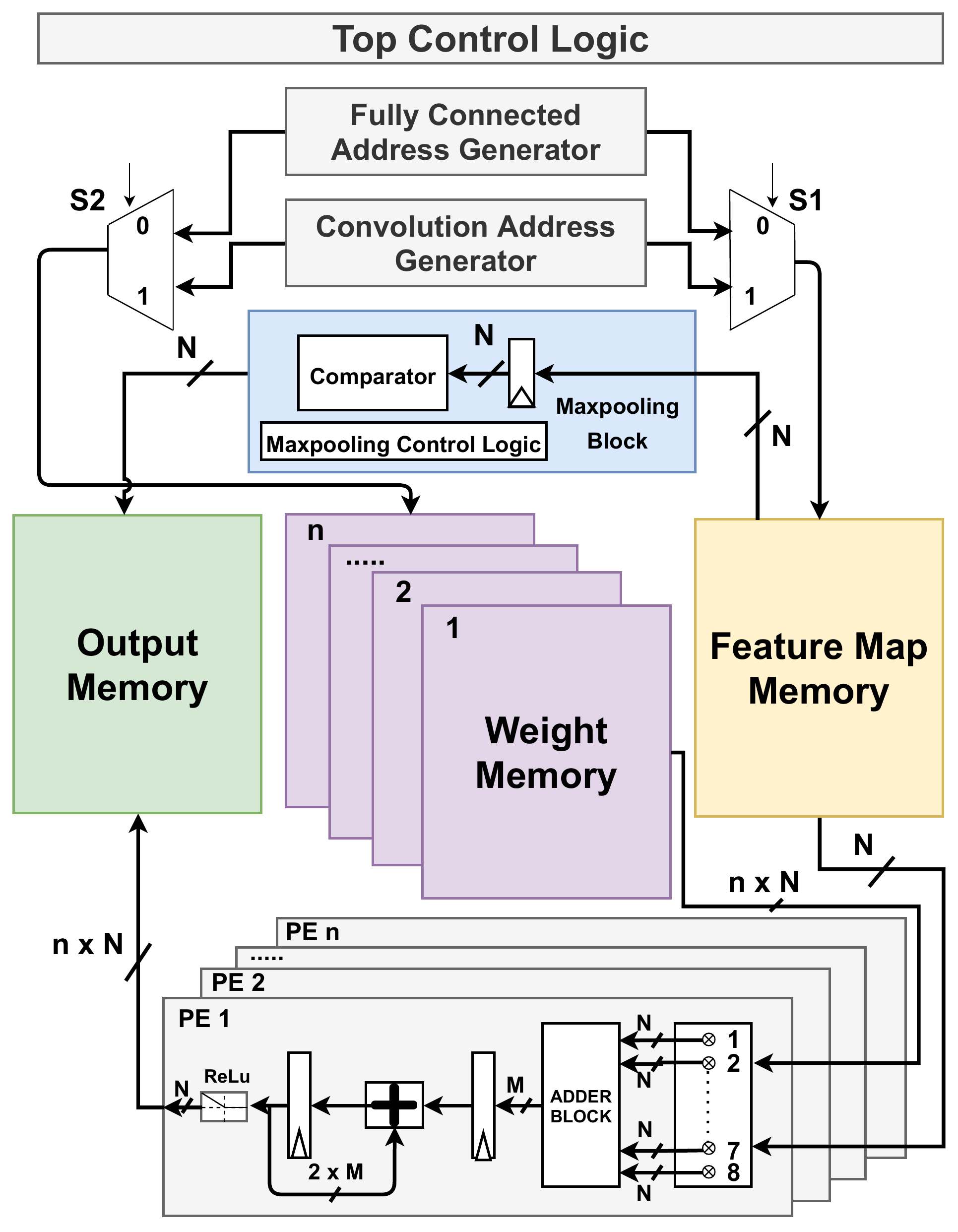}
\end{center}
\vspace{-3ex}
\caption{The high-level overview of the proposed accelerator design. The feature map memory forwards input data to the PE array while the weight memory alternates layer weights for computations inside the PE array. The output from the MAC operation is temporarily stored in the output memory. Along with this, the maxpooling block performs maxpooling function of the data with the help of a comparator where it bubble sorts the data to achieve proper feature map size. Top control logic regulates the state machine and pipelines the order of execution for convolution, maxpooling, and fully connected layers for precise hardware operation.}
\vspace{-3ex}
\label{fig:top_hard}
\end{figure}

The PE array houses $P$ processing engines, each equipped with 8 MAC units to replicate the activation operation of convolution and allow parallel processing. The memories include a feature map and output memory to store the input and output data. Similarly, the weights are stored in the weight memory. The width of each memory depends on the number of multipliers $M$ and is defined by $Mq$. On the other hand, the depth is characterized by $Largest\_feature\_map\_size/P/M/q$, for feature map and output memory and $Model\_size/P/M/q$ for weights, respectively. The convolution tiling technique utilizes the output channel tiling process where each PE only operates on one output channel. For our design, the peak performance of the design is designated by $2.min(F, P).min(C, M).frequency$ where $F$ and $C$ correspond to the number of filters or output channels and input channels respectively, and $frequency$ is the operating frequency of the CNN hardware. Fully connected (FC) operation is implemented through the same pipeline that is used for convolution. However, maxpooling requires a comparator to employ the bubble sort strategy that iterates through the data to select the maximum pixel value. We choose the value for $P$ and $M$ based on the intuition that we want to get the most performance out of our implementation while also ensuring that none of the PEs are idle during the execution of any of the convolution layers to avoid underutilization of resources. Throughout all the experiments, the minimum number of channels in our neural network turned out to be 16; hence, we selected $PE$ as 16$s$. To simplify our computations, we limit $M$ to only 8 multipliers.

\section{Energy Regression}
\label{sec:ene_reg}

In order to perform energy regression, first, we need to come up with a way to configure the performance, latency, and power for our design. The performance and latency for our method can be defined by Equation \ref{eq:perf&latency}.

\begin{equation} 
\footnotesize
\begin{aligned}
% \begin{split}
{Performance}_{_{FC/CONV}}(\,\mathcal{HW}_{_{efficient}}\,|\,\mathcal{NN}\scriptstyle\langle q,\,s\rangle \displaystyle\,) &\propto s+\frac{s}{8}\\
\qquad {Latency}_{_{FC/CONV}}(\,\mathcal{HW}_{_{efficient}}\,|\,\mathcal{NN}\scriptstyle\langle q,\,s\rangle \displaystyle\,) &\propto s+C 
% \end{split}
\end{aligned}
\label{eq:perf&latency}
\end{equation}

The first term in the performance equation of Equation \ref{eq:perf&latency} corresponds to the performance of all layers except the first one. The performance of the first layer is different as it only has one input channel and is thus signified by the second term in the Equation \ref{eq:perf&latency}. This suggests that when the hardware is scaled by $s$, both the total number of computations and processing engines in our hardware for all layers involved are scaled by $(s+s/8)$ where $s/8$ is a constant accounting for the performance of the first layer. The factor of 8 here denotes that for the first layer, the performance is always scaled by the minimum of 8 and 1 $[min(8,1)]$ where 8 is the number of multipliers in the design and 1 is the number of channels in the first layer. The equation for latency comes from considering both the total number of computations in the network and the performance of all the layers, which results in $(s+C)$ where $C$ represents the latency for the first layer. The final latency numbers and performance of a number of our experiments with varying scaling and bit precision are tabulated in Table \ref{tab:summary}.

We estimate that for our accelerator design, the majority of the power dissipation comes from four distinct elements in the design, which are (1) memory communication, (2) multiplication, (3) addition, and (4) static power of the device. With regards to this, we formulate that power can be represented by Equation \ref{eq:power_sq} where power coming from the memory is proportional to $Mq$, the power coming from the multipliers and adders are proportional to $PMq^2$ and $PMq$, respectively. $\hat{B_0}$, in this case, denotes the static power.

\begin{equation} 
\footnotesize
\begin{aligned}
% \begin{split}
Power(\mathcal{HW}|\mathcal{NN}\scriptstyle\langle q,\,s\rangle \displaystyle) &&\approx \hat{B_3}.q^2.s^2+\hat{B_2}.q.s^2+\hat{B_1}.q.s+\hat{B_0}
% \end{split}
\end{aligned}
\label{eq:power_sq}
\end{equation}

In our experiments, the latency changes in proportion to the scaling, directly affecting the experimented workloads' energy consumption. Using the power equation from Equation \ref{eq:power_sq} and latency equation from Equation \ref{eq:perf&latency}, we model the energy as the following:

\begin{equation} 
\footnotesize
\begin{aligned}
% \begin{split}
Energy(\mathcal{HW}|\mathcal{NN}\scriptstyle\langle q,\,s\rangle \displaystyle) &\approx (\hat{B_3}.q^2.s^2+\hat{B_2}.q.s^2+\hat{B_1}.q.s+\hat{B_0})(\hat{D}.s+\hat{E})
% \end{split}
\end{aligned}
\label{eq:energy_sq}
\end{equation}

\begin{figure}[]
\begin{center}
\includegraphics[width=\linewidth]{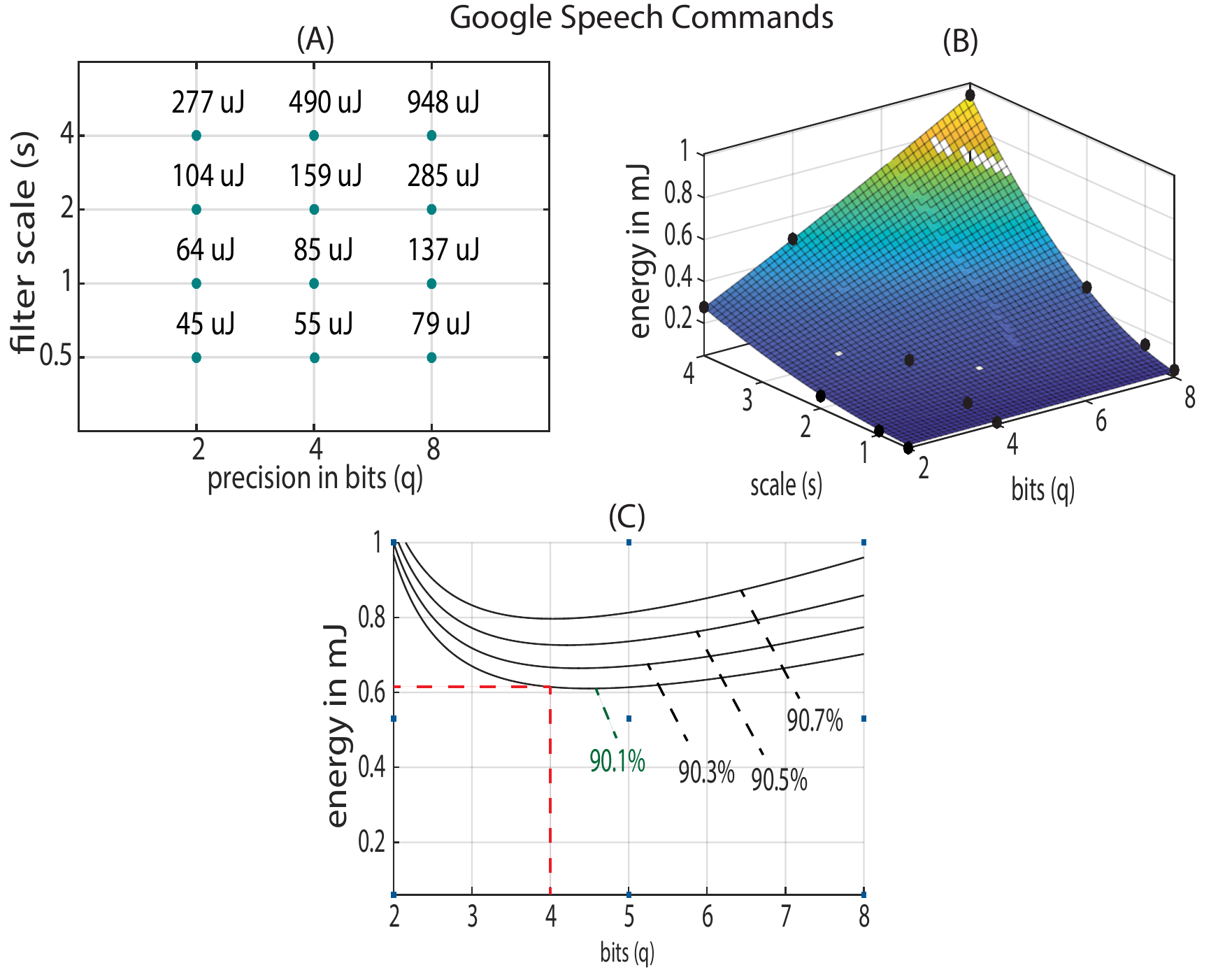}
\end{center}
\vspace{-3ex}
\caption{Energy modeling setup based on experimental analysis for Google speech commands. (A) represents the energy consumption experienced from implementing different $\mathcal{NN}$ configurations, (B) illustrates the surface plot indicating the fit for the energy equation, and (C) demonstrates a better visualization of the relationship between accuracy, energy, and precision through contours.}
\vspace{-3ex}
\label{fig:energy_modeling}
\end{figure}

Here, $\hat{B_0}$, $\hat{B_1}$, $\hat{B_2}$, $\hat{B_3}$, $\hat{D}$, and $\hat{E}$ are learnable parameters. The terms in the first parenthesis of Equation \ref{eq:energy_sq} relate to power from multiplications, power from additions, power from communication between memories, and static power of the device. Again, the terms in the second parenthesis outline the latency coming from all the layers. To generate data for our regression, we attempt to have at least a 12-point setup where the filter scale varies from 0.5 up to 4 and precision changes from 2 up to 8-bits. The fitted surface plots from this 12-point regression system are shown in Figure \ref{fig:energy_modeling}. The RMSE for the energy estimator turns out to be 0.01 mJ, indicating a very good fit.

From surface plots of energy against scale and quantization, we can further generate energy contour plots shown in Figure \ref{fig:energy_modeling} (C). To achieve this, we take $s$ as a function of $q$ and $Accuracy$ to substitute in the $Energy$ equation. This gives us expected convex curves that highlight energy consumption against quantization bits for different levels of accuracy. To ensure hardware-friendly specifications, we choose $q$ such that it is a natural number around the minima of the convex curves and $s$ such that $16s$ is a natural number. This also allows us to profile the energy consumption for unknown combinations of $q$ and $s$ and thereby assist in selecting the configuration which has the best metrics (accuracy and energy) from both architecture and deployment perspectives.

\section{Implementation Results}
\label{sec:analysis}

The complete implementation results of a few configurations are tabulated in Table \ref{tab:summary}. It was evident from the accuracy regression contour in Figure \ref{fig:accuracy_modeling} that we can only get very high accuracy levels (around 90\%) with a network implemented with 4-bits and above. Hence, we only demonstrate the hardware experiments for 4-bit precision and above for varying scaling. It is also important to note that energy efficiency considerably increases as the networks become more scaled. With higher scaled networks, we need to allocate more PEs to work on the increased output channel, thereby ensuring minimal provision for any PE to be inactive. The quality of fit for accuracy and energy regression demonstrated by Figure \ref{fig:accuracy_modeling} and Figure \ref{fig:energy_modeling} respectively is also verified through estimation in Table \ref{tab:verify}. We test on new data points ($\mathcal{NN}$ $\scriptstyle\langle q=4,\,s=2.5\rangle \displaystyle$, $\mathcal{NN}$ $\scriptstyle\langle q=4,\,s=4.5\rangle \displaystyle$) both experimentally and through our optimal regression equations. For both accuracy and regression, there is very little deviation between the actual values and the predicted values as per Table \ref{tab:verify} which ensures our regression is a good fit. From Figure \ref{fig:energy_modeling} we also observe that we can get 90.1\% accuracy with both a 4-bit and 5-bit implementation with approximately similar energy cost. We can choose any of the 4-bit or 5-bit precision for the scaling of 4.5 in this case. Though for our implementation, we used the simpler implementation of 4-bits, 5-bit implementation of the same scaling is equally viable. Figure \ref{fig:energy_modeling} also informs us that to have around 90\% accuracy with lower precision configurations(2-bits and 3-bits), we would have to scale up the networks considerably, thus resulting in more energy consumption. On the other hand, we can use 8-bit networks with lower scaling, but the overall energy consumption turns out to be higher than 4-bit and 5-bit representations. 

\begin{table}[]
\centering
\caption{Implementation Results of Different Workloads on the AC~701 Platform at 100 MHz}
\label{tab:summary}
\begin{tabular}{c|c|c|c|c|c}
\hline
($q,s$) & ($P,M$) & BRAM & Pwr(W) & GOPJ & Latency (ms) \\ \hline
(4.0,1.0) & (16,8) & 18 & 0.28 & 40.1 & 0.31 \\ \hline
(4.0,2.0) & (32,8) & 51 & 0.38 & 79.6 & 0.42 \\ \hline
(4.0,4.0) & (64,8) & 165 & 0.76 & 98.9 & 0.65 \\ \hline
(4.0,4.5) & (72,8) & 185 & 0.83 & 104.9 & 0.70 \\ \hline
(8.0,1.0) & (16,8) & 27 & 0.45 & 24.9 & 0.31 \\ \hline
(8.0,2.0) & (32,8) & 85 & 0.68 & 44.5 & 0.42 \\ \hline
(8.0,4.0) & (64,8) & 296 & 1.47 & 51.1 & 0.65 \\ \hline
\end{tabular}
\end{table}

\begin{table}[]
\centering
\caption{Verification of Energy and Accuracy Regression with New Datapoints}
\label{tab:verify}
\begin{tabular}{c|c|cc|cc}
\hline
\multirow{2}{*}{($q,s$)} & \multirow{2}{*}{($P,M$)} & \multicolumn{2}{c|}{Energy (mJ)} & \multicolumn{2}{c}{Accuracy(\%)} \\
 &  & Actual & Pred. & Actual & Pred. \\ \hline
(4.0,2.5) & (40,8) & 0.21 & 0.23 & 86.5 & 86.4 \\ \hline
(4.0,3.5) & (56,8) & 0.39 & 0.41 & 88.7 & 84.3 \\ \hline
(4.0,4.5) & (72,8) & 0.58 & 0.60 & 90.3 & 90.1 \\ \hline
(8.0,2.5) & (40,8) & 0.41 & 0.42 & 88.7 & 88.5 \\ \hline
(8.0,3.0) & (48,8) & 0.56 & 0.59 & 89.6 & 89.4 \\ \hline
(8.0,3.5) & (56,8) & 0.74 & 0.76 & 90.2 & 90.0 \\ \hline
\end{tabular}
\end{table}

\section{Comparison with Existing Works}
\label{sec:comparison}

We compare the performance of our accelerator implementation with two recent works that focus on accelerating neural networks for KWS on resource-constrained hardware. \cite{dinelli2019fpga} uses an accelerator framework developed on Xilinx Artix-7 FPGA for KWS on the speech commands dataset. This implementation concentrates on quantizing data to 8-bit fixed points for reducing the memory footprint with separable convolution instead of vanilla convolution layers. It also considers the MFCC coefficient matrix as input like our processing. Another work on \cite{zhang2017hello} accelerates a traditional CNN-based architecture on the Cortex-M7 microcontroller. Microcontrollers represent tiny resource-constrained platforms and usually allow fast inference and a low power envelope. In order to make an appropriate comparison to these works, we accelerated our network with a scaling of 4.5 and a quantization level of 4-bits that provided us with the best trade-off for accuracy against energy as per Figure \ref{fig:energy_modeling} (C). We also implemented a variation of our network to be run at 47.6 MHz to be on the same frequency level as \cite{dinelli2019fpga}. Compared to \cite{dinelli2019fpga} our deployment has almost 4.5$\times$ the workload with respect to model size but dissipates slightly lower power with similar latency and higher accuracy. In terms of model size, we significantly reduce memory requirements by choosing a 4-bit implementation compared to \cite{zhang2017hello} and this configuration allows our design to have an energy and energy efficiency advantage of around 2.1$\times$ and 4$\times$ respectively compared to \cite{dinelli2019fpga}. On the other hand, when run at the standard 100 MHz frequency, our design is approximately 8.5$\times$ faster than the microcontroller implementation in \cite{zhang2017hello}. The primary reason behind the improvements is that our parallelization technique employs a form of output channel tiling where the majority of the PEs are active, thus ensuring that the performance is very close to the peak performance of the design. Finally, our network is small enough to be accelerated through a small microcontroller or low-cost edge device with a model size of only 340~KB. 

\begin{table}[]
\centering
\caption{Comparison of our $\mathcal{NN}$ $\scriptstyle\langle q=4,\,s=4.5\rangle \displaystyle$ implementation on the XC7A200T platform to recent keyword spotting hardware implementations.}
\label{tab:comparison}
\begin{tabular}{c|c|c|cc}
\hline
Related work & \cite{dinelli2019fpga} & \cite{zhang2017hello} & \multicolumn{2}{c}{This Work} \\ \hline
Model & SCNN & CNN & \multicolumn{2}{c}{CNN and FC} \\ \hline
Dataset & \begin{tabular}[c]{@{}c@{}}Speech \\ Commands\end{tabular} & \begin{tabular}[c]{@{}c@{}}Speech \\ Commands\end{tabular} & \multicolumn{2}{c}{\begin{tabular}[c]{@{}c@{}}Speech\\ Commands\end{tabular}} \\ \hline
Precision & 8-bits & 8-bits & \multicolumn{2}{c}{4-bits} \\
Accuracy(\%) & 88.1 & 87.6 & \multicolumn{2}{c}{90.1} \\ \hline
Device & XC7A200T & \begin{tabular}[c]{@{}c@{}}Cortex-M7 \\ Microcontroller\end{tabular} & \multicolumn{2}{c}{XC7A200T} \\ \hline
Model Size (KB) & 150 & 497 & \multicolumn{2}{c}{340} \\ \hline
Power (W) & 1.04 & - & \multicolumn{1}{c|}{0.47} & 0.83 \\
Clock (MHz) & 47.6 & 216 & \multicolumn{1}{c|}{47.6} & 100 \\
Latency (ms) & 1.43 & 12 & \multicolumn{1}{c|}{1.47} & 0.70 \\
Energy (mJ) & 1.49 & - & \multicolumn{1}{c|}{0.70} & 0.58 \\
GOPJ & 10.5 & - & \multicolumn{1}{c|}{41.5} & 104.9 \\ \hline
\end{tabular}
\end{table}

\section{Conclusion}
\label{sec:conclusion}

KWS is widely used in voice assistant devices nowadays. Due to the requirement of KWS systems to be always on, low power consumption and energy-efficient representations of the KWS networks are of primary importance. To this end, we propose a regression-oriented network exploration setup that finds out the best configurations, leading to desired accuracy specifications and an appropriate accelerator design in a prompt fashion. In this work, we only use a 12-point regression procedure, the values of which can be easily generated by training a handful of neural networks in parallel. Additionally, this strategy can also be utilized on top of different optimization techniques to allow further flexibility to the optimization solutions. A future extension of this work can include introducing this setup to minimize the latency and energy consumption of microcontroller representations for KWS.

\bibliographystyle{ACM-Reference-Format}
\bibliography{ref.bib, eehpc.bib}

\end{document}